%% file: cvpr.tex
\documentclass[final]{cvpr}

\usepackage{times}
\usepackage{epsfig}
\usepackage{graphicx}
\usepackage{amsmath}
\usepackage{amssymb}

\usepackage{algorithm}
\usepackage[noend]{algpseudocode}
\usepackage{setspace}

\usepackage{amsfonts}
\usepackage{amsmath}
\usepackage{mathtools}

\usepackage{booktabs}       
\usepackage{wrapfig}

\usepackage{color}
\usepackage[usenames,dvipsnames,table]{xcolor}
\definecolor{DarkCoral}{rgb}{0.8, 0.36, 0.27}
\definecolor{darkpastelgreen}{rgb}{0.01, 0.75, 0.24}
\definecolor{ao(english)}{rgb}{0.0, 0.5, 0.0}
\definecolor{forestgreen(web)}{rgb}{0.13, 0.55, 0.13}
\definecolor{green(pigment)}{rgb}{0.0, 0.65, 0.31}

\definecolor{brickred}{rgb}{0.8, 0.25, 0.33}

\definecolor{blue(ncs)}{rgb}{0.0, 0.53, 0.74}

\newcommand{\metada}{\textit{Meta-DR}}

\newcommand{\myparagraph}[1]{\vspace{0.1cm}\noindent\textbf{#1.}}

\usepackage[pagebackref=true,breaklinks=true,colorlinks,bookmarks=false]{hyperref}

\def\cvprPaperID{8086}
\def\confYear{CVPR 2021}

\begin{document}

\title{Continual Adaptation of Visual Representations\\ via Domain Randomization and Meta-learning}

\author{

Riccardo Volpi\\
\and
Diane Larlus\\~\\
NAVER LABS Europe\thanks{\href{www.europe.naverlabs.com}{www.europe.naverlabs.com}}\\
{\tt\small \{name.lastname\}@naverlabs.com}

\and
Gr\'egory Rogez\\

}
\maketitle

\input{0_abstract}

\input{1_intro}

\input{2_relwork}
\input{3_probform}
\input{4_method}
\input{5_experiments}
\input{6_conclusions}

\myparagraph{Acknowledgments}
This work is part of MIAI@Grenoble Alpes (ANR-19-P3IA-0003).

\clearpage

{\small
\bibliographystyle{plain}
\bibliography{egbib}
}

\appendix
\input{supplementary}

\end{document}

%% file: 0_abstract.tex
\begin{abstract}
Most standard learning approaches lead to fragile models which are prone to
drift when sequentially 
trained on samples of a different nature---the well-known \textit{catastrophic
forgetting} issue. 
In particular, when a model consecutively learns from different visual domains, it tends to forget the past domains in favor of the most recent ones. In this context, we show that one way to learn models that are inherently more robust against forgetting is domain randomization---for vision tasks, randomizing the current domain's distribution with heavy image manipulations.
Building on this result, we devise a meta-learning strategy where a regularizer explicitly penalizes
any loss associated with transferring the model from the current domain to different
``auxiliary'' meta-domains, while also easing adaptation to them. Such meta-domains are also generated
through randomized image manipulations.
We empirically demonstrate in a variety of experiments---spanning from classification to semantic segmentation---that our
approach results in models that are less prone to catastrophic
forgetting when transferred to new domains.
\end{abstract}

%% file: 1_intro.tex
\section{Introduction}\label{sec:intro}

Modern computer vision approaches can reach super-human performance in a
variety of well-defined and isolated tasks at the expense of versatility. When confronted to 
a plurality of new tasks or new visual domains, they have trouble adapting, or adapt at the cost
of forgetting what they had been initially trained for. This phenomenon, that has been observed
for decades~\cite{mccloskey:catastrophic}, is known as
\textit{catastrophic forgetting}.
Directly tackling this issue, \textit{lifelong learning} approaches, also known as
\textit{continual learning} approaches, 
are designed to continuously learn from new samples without forgetting the past. 

\begin{figure}
    \includegraphics[width=\linewidth]{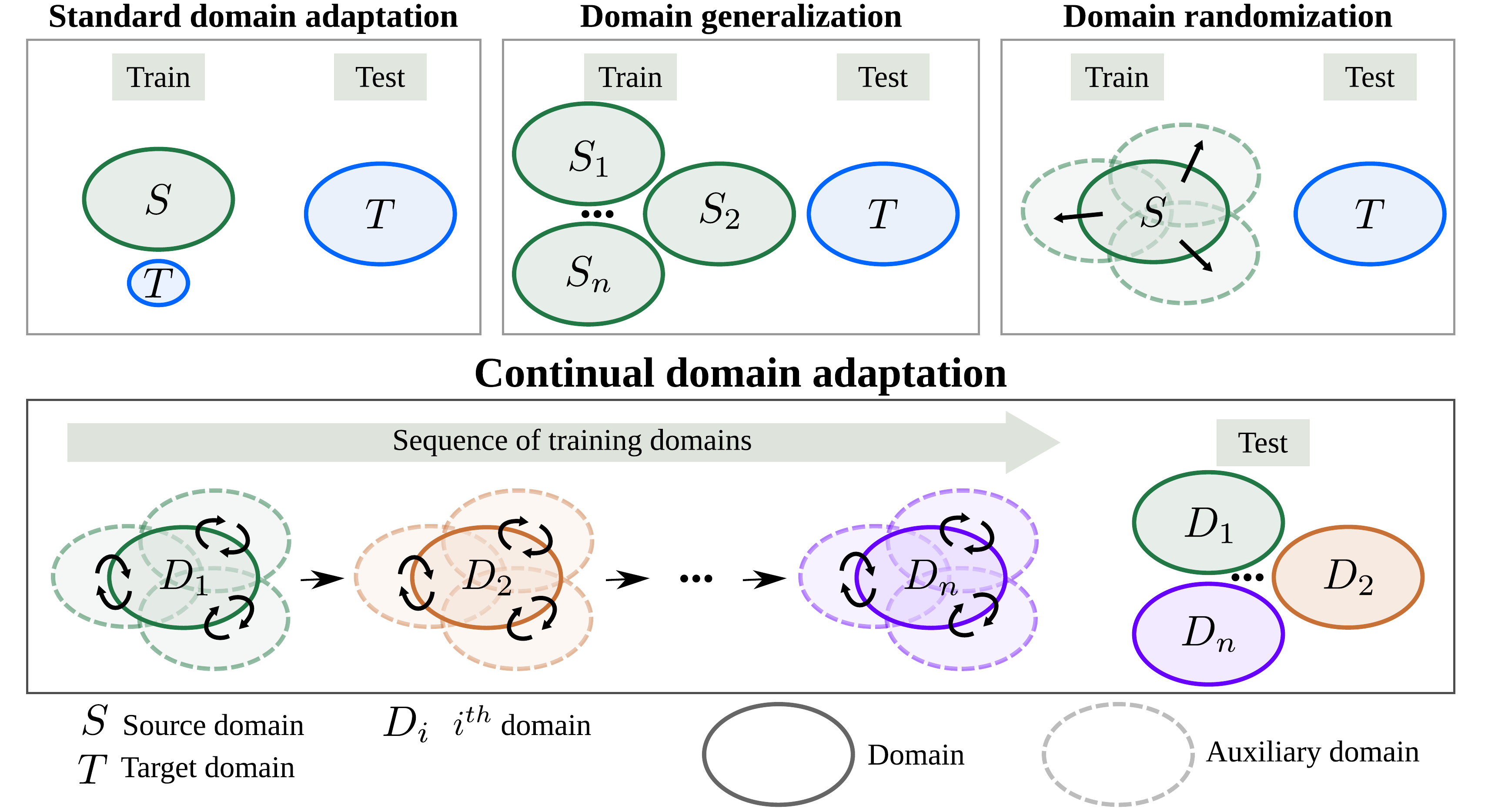}
    \caption{This paper tackles the \textbf{continual domain adaptation} task \textit{(bottom)},
    compared here with standard domain adaptation, domain generalization, and domain randomization
    \textit{(top)}.}
    \label{fig:continual_DA}
    \vspace{-15pt}
\end{figure}

In this work, we are concerned with the problem of {\em continual} and {\em supervised} adaptation to new
visual domains (see Figure~\ref{fig:continual_DA}). 
Framing this task as \textit{continual domain adaptation}, we 
assume that a model has to learn to perform a given task while being exposed to conditions which constantly change 
throughout its lifespan. This is of particular interest when deploying applications 
to the real-world where a model is expected to seamlessly adapt to its
environment and can encounter different domains from the one(s) observed originally at
training time.
A possible solution to mitigate catastrophic forgetting is storing samples from all
domains encountered throughout the lifespan of the model~\cite{parisi2019continual}. While
effective,
this solution may not be appropriate when retaining data is not allowed 
(\textit{e.g.}, due to privacy concerns) or when working under strong memory
constraints (\textit{e.g.}, in mobile applications).
For these reasons, we are interested in developing methods for learning visual
representations that are 
inherently more robust against catastrophic forgetting, without necessarily requiring any
information storage or model expansion.

To tackle this problem, we start from a simple intuition: when adapting a model trained on a
domain $\mathcal{D}_1$ to a second domain $\mathcal{D}_2$, we can anticipate the severity of
forgetting to depend on how demanding the adaptation process is---that is, how close $\mathcal{D}_2$ is to $\mathcal{D}_1$.  
The natural issue is that we cannot control whether sequential domains
will be more or less similar to each other. Motivated by results on domain
randomization~\cite{tobin2017domainrandomization,yue2019domainranzomization} and single-source
domain generalization~\cite{volpi2019model}, we propose to heavily perturb the distribution of
the current domain to increase the probability that samples from future domains will be closer to the current data distribution---hence (generally) requiring a lighter adaptation
process. Focusing on computer vision tasks, we use image transformations for the randomization
process, and show that models trained in this fashion are significantly more robust against catastrophic forgetting in the context of continual,
supervised domain adaptation. This result represents our first contribution.

Further, we question whether we can learn representations that are \textit{inherently} robust
against transfer to new domains (that is, against gradient updates on samples from distributions
different than the current one).
We tackle this problem through the lens of meta-learning and devise a
regularization strategy that forces the model to train for the task of interest
on the current domain, while learning to be resilient to potential parameter updates on domains
different from the current one. In general, meta-learning approaches require access to a number
of different meta-tasks (or \textit{meta-domains} in our case), but
our setting only allows access to samples from the {\em current domain} at any point in time.
To overcome this issue, we introduce ``auxiliary'' meta-domains that are
produced by 
randomizing the original distribution---also here, using standard image transformations.
Additionally, inspired by 
Finn et al.~\cite{finn2017model}, we encourage our model to train in a way that will allow it to
efficiently adapt to new domains. The devised meta-learning algorithm, based on the new concept of \textit{auxiliary meta-domains}, constitutes our second contribution.

To extensively assess the effectiveness of our continual domain adaptation strategies, we start with an experimental protocol which tackles digit recognition.
Further, we increase the difficulty of the task and focus on the PACS dataset~\cite{li2017deeper} from the domain generalization literature, used here to define learning trajectories across different visual domains.
Finally, we focus on semantic segmentation, exploring learning sequences across different simulated urban environments and weather conditions.
In all the aforementioned experiments, we show the benefits of the proposed approaches. To conclude our analysis, we show that our methods can further be improved by combining them with a small memory of samples from previous domains~\cite{chaudhry2019tiny}.

%% file: 2_relwork.tex
\section{Related work}\label{sec:relwork}

Our work lies at the intersection of lifelong learning, data augmentation, meta-learning, and
domain adaptation. 
We provide here an abridged overview of the relevant background
and refer the reader to Parisi~\etal~\cite{parisi2019continual}, Hospedales~\etal~\cite{hospedales2020metalearning} and Csurka~\cite{csurka17domain} for  detailed reviews on lifelong learning, meta-learning and domain adaptation. 

\myparagraph{Lifelong learning}
The main goal of lifelong learning research is devising models that can learn new information throughout their lifespan, without forgetting old patterns. The survey of Parisi \etal~\cite{parisi2019continual} divides lifelong learning approaches
into three categories:
(i) dynamic architectures, where the model's
underlying architecture is modified as it learns
new
patterns~\cite{rusu2016progressive,zhou2012online,cortes2017adanet,xiao2014error,draelos2017neurogenesis,yoon2018lifelong,rebuffi2017icarl}; (ii)
rehearsal methods, that rely on memory replay and overcome catastrophic forgetting by storing samples from old
tasks/distributions and regularly feeding them again to the model~\cite{hinton1987using,lopexpaz2017nips,chaudry2019efficient,chaudry2019continual,riemer2019learning,hayes2020remind,prabhu2020gdumb};
(iii) regularization
methods, that propose ways of constraining the tasks'
objectives in order to avoid
forgetting~\cite{li2016learning,kirkpatrick2017overcoming,zenke2017continual,fini2020online}.
Our work is of the third flavor: we tackle lifelong learning
without necessarily replaying old data nor increasing the
model capacity over time---even though we also show that our methods can be used in tandem with a small memory of old samples. 
In contrast with most or the prior art that focuses on task/class incremental learning, we tackle scenarios where the domain sequentially changes but the task remains the same; we refer to this problem as continual domain adaptation. See Van de Ven and Tolias for a closer look at the different problem formulations~\cite{van2019three}.

\myparagraph{Data augmentation and domain randomization}
The use of data
augmentation has a long history in computer
vision~\cite{scholkopf1996incorporating,simard2003bestpractices,ciresan2011highperformance,ciresan2012multicolumn,krizhevsky2012imagenet,RogezS16}.
Applying geometric or photometric transformations to the images of the training set generates
new training samples for free and constitutes an effective
strategy to improve performance. 
Randomizing the input distribution has been shown to be particularly effective to improve
sim-to-real performance (domain
randomization~\cite{tobin2017domainrandomization,yue2019domainranzomization}), and also to
improve out-of-domain performance in single-source domain generalization
problems~\cite{volpi2019model} (see
Figure~\ref{fig:continual_DA}, top).
In this work, we first show that randomizing the domain at hand with heavy image manipulations
helps preventing catastrophic
forgetting in a continual domain adaptation setting. Then, we leverage similar
transformations to automatically generate the samples that compose our ``auxiliary''
meta-domains. Our experimental results show that the proposed
meta-learning strategy improves over simply using these additional samples in a standard
data augmentation fashion.\\

\begin{figure*}[t!]
    \begin{center}
        \includegraphics[width=1\textwidth]{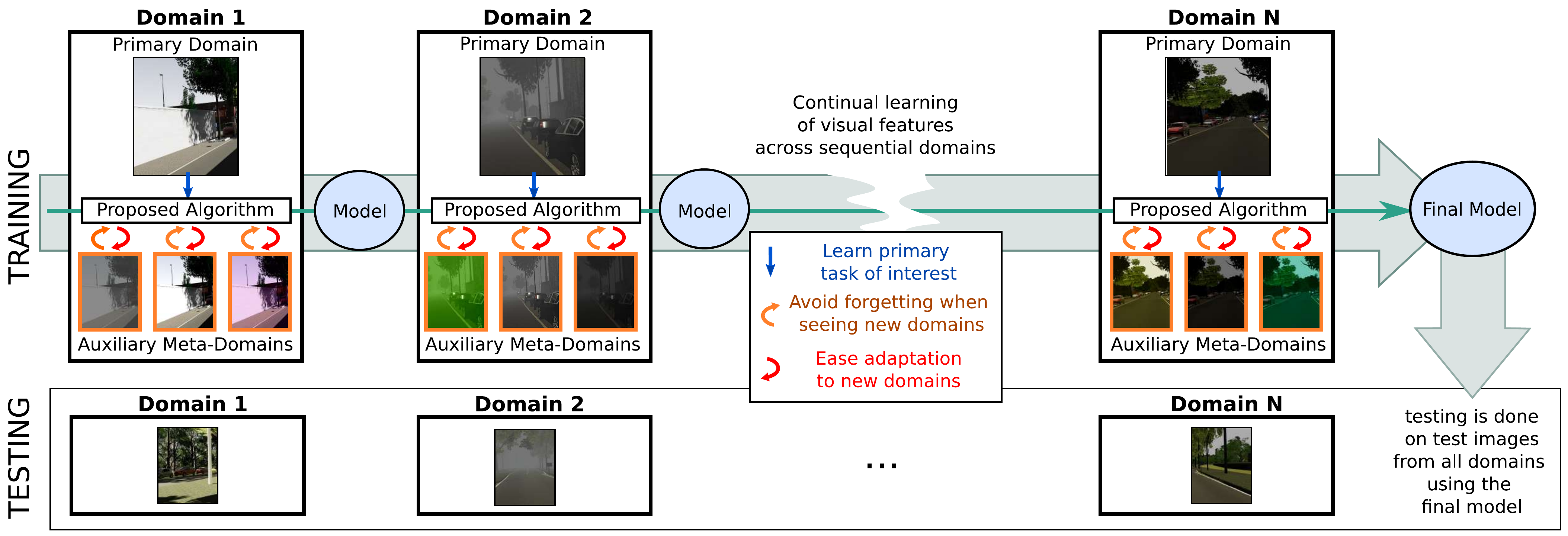}
    \end{center}
 \vspace{-5pt}
\caption{
Life-cycle of a model when training for continual domain adaptation. At every new encountered domain, our proposed method (Algorithm~\ref{alg:metatrain}: \metada) is applied on the training set of that domain (Primary Domain) and on the generated ``auxiliary'' Meta-Domains. The final model is evaluated on test images from all the encountered domains to evaluate resilience to catastrophic forgetting.}
\vspace{-10pt}
\label{fig:main}
\end{figure*}

\myparagraph{Meta-learning}
We draw inspiration from
meta-learning approaches~\cite{li2017learning,andrychowicz2016learningtolearn,ravi2017optimization},
and in particular from the idea of learning representations that can be efficiently transferred to new tasks~\cite{finn2017model}.
Meta-learning generally relies on a series of
meta-train and meta-test splits, and the optimization process enforces that a
few gradient descent steps on the meta-train splits lead to good
generalization performance on the meta-test.
Similar to this approach, we devise a two-fold regularizer: on
the one hand, it encourages models to remember previously encountered domains when exposed to
new ones (by means of gradient descent updates on these tasks); on the
other hand, it also encourages an efficient adaptation to such
domains. In contrast with related work that relies on meta-learning to handle 
continual learning problems, we do not require access to
a buffer of old samples~\cite{riemer2019learning}, nor focus on learning from data streams~\cite{javed2019metalearning}.\\

\myparagraph{Domain adaptation}
Focusing on the resilience to domain-shifts rather than the more
standard task-shift, our work can also be put in the context
of the domain adaptation literature~\cite{shimodaira00improving,ben-david07analysis,daume06domain,saenko2010adapting,ganin2015unsupervised}.
Our aim is indeed to perform \textit{continual domain adaptation}, without
degrading performances on past domains while facing new ones.
Yet, there is a fundamental difference with the premises of our work.
While, in the standard domain adaptation literature, the sole role of the
source domain(s) is to compensate for the scarcity of annotated data from
the target domain, in either supervised or unsupervised
settings~\cite{csurka17domain}, our formulation does not assume such scarcity
and cares about performance on the entire sequence of encountered domains.

%% file: 3_probform.tex
\section{Notations and problem formulation}\label{sec:probform}

First, we formally introduce the notions of catastrophic forgetting and continual
domain adaptation. Then, we formalize the problem tackled in this paper and
introduce the baseline we build upon. 

\myparagraph{Notations and definitions}
Let us assume that we are interested in training a model $\mathcal{M}_{\theta}$ to solve a task $\mathcal{T}_0$, relying on some
data points that follow a distribution $\mathcal{P}_0$. In practice, we usually do not know this distribution 
but are provided with a set of samples $\mathcal{S}_0\sim\mathcal{P}_0$. We focus on
supervised learning 
and assume $m$ training samples 
$\mathcal{S}_0 = \{(\mathbf{x}_{k},\mathbf{y}_{k})\}_{k=1}^{m}$, where $\mathbf{x}_k$ and $\mathbf{y}_k$
represent a data sample and its corresponding
label, respectively.
We train our model by empirical risk minimization (ERM), optimizing a loss
$\mathcal{L}_{\mathcal{T}_0}(\theta)$. For example, for the supervised training of a multi-class classifier, this loss is typically the cross-entropy 
between the predictions of the model $\hat{\mathbf{y}}$ and the ground-truth annotations $\mathbf{y}$:

\vspace{-10pt}
{
\small
\begin{equation}\label{eq:erm}
\theta^*_{\mathcal{T}_0} = \min_{\theta}\Big\{\mathcal{L}_{\mathcal{T}_0}(\mathcal{S}_0; \theta) \coloneqq -\frac{1}{m}\sum_{k=1}^m \mathbf{y}_k^{T}\log\hat{\mathbf{y}}_k\Big\}.
\vspace{-10pt}
\end{equation}
}

While neural network models trained via ERM (carried out via gradient descent) have been effective in a broad range of problems, they are prone to forget about
their initial task
when fine-tuned on a new one, even if the two tasks appear very similar at first glance.

In practice, this means that if we use the model  $\theta^*_{\mathcal{T}_0}$
trained on the first task $\mathcal{T}_0$ as 
a starting point to train for a different task 
$\mathcal{T}_1$,
the newly obtained model $\theta^*_{\mathcal{T}_0 \rightarrow \mathcal{T}_1}$ 
typically shows degraded performances on $\mathcal{T}_0$. More formally,
$\mathcal{L}_{\mathcal{T}_0}(\theta^*_{\mathcal{T}_0 \rightarrow \mathcal{T}_1}) >
\mathcal{L}_{\mathcal{T}_0}(\theta^*_{\mathcal{T}_0})$.
This undesirable property of deteriorating performance on the previously
learned task is known as \textit{catastrophic forgetting}. 

\myparagraph{Problem formulation}
In this work, we assume that the task remains the same but that the domain
varies instead. The model is sequentially
exposed to a list of different domains. We wish for the model to being able to adapt
to each new domain without degrading its performance on the old ones. We refer to this as \textit{continual
  domain adaptation}.
The goal of this paper is to mitigate catastrophic forgetting on previously seen
domains.

More formally, given a task $\mathcal{T}$ that remains
constant, we assume that the model is exposed to a sequence of $N$ domains $(D_i)_{i=1}^N$,
each characterized by a distribution $\mathcal{P}_i$ from which specific samples
$\mathcal{S}_i$ are drawn.
As in previous work~\cite{lopexpaz2017nips}, we assume locally i.i.d. data distributions.
With this new formulation, and slightly abusing the notations, the problem of catastrophic forgetting mentioned
before can be rewritten as 
$\mathcal{L}_{\mathcal{T}}(\theta^*_{D_i \rightarrow D_{i+1}}) >
\mathcal{L}_{\mathcal{T}}(\theta^*_{D_i}).$
We assume that each set of samples $\mathcal{S}_i$
becomes unavailable when the next domain $D_{i+1}$ with samples $\mathcal{S}_{i+1}$ is
encountered. 
We are interested in assessing the performance of the model 
at the end of the training sequence, and for every domain $D_i$ (see Figure~\ref{fig:main}).

\myparagraph{Baseline}
A naive approach to tackle the problem defined above is simply to start from the model
$\mathcal{M}_\theta^i$ obtained after training on domain $D_i$ and to fine-tune
it using samples from $D_{i+1}$. Due to catastrophic forgetting, this baseline typically
performs poorly on older domains $i < N$ when it reaches the end of its training cycle.
This will be our experimental lower-bound.

%% file: 4_method.tex
\section{Method}\label{sec:method}

\subsection{Image transformation sets}
\label{sec:da}

A core part of our study is the definition of proper image transformation sets, used to drive the domain randomization process---and, for what concerns the proposed meta-learning solution, to define the auxiliary meta-domains.
We assume access to a set $\Psi$,
where each element is a specific transformation with a specific magnitude level (\textit{e.g.},
``increase brightness by $10\%$''). Drawing from Volpi and
Murino~\cite{volpi2019model}, we consider a larger set covering all the possible
transformations obtained by combining $N$ given basic ones
(\textit{e.g.}, with $N=2$, ``increase brightness by $10\%$ and then reduce contrast by $5\%$''). 

Given the so-defined set and a sample set $\mathcal{S}=
\{(\mathbf{x}_{k},\mathbf{y}_{k})\}_{k=1}^{m}\sim\mathcal{P}$, we can craft novel data points by
sampling an object from the set 
$T\sim\Psi$, and then applying it to the given data points,
obtaining 
$\mathcal{\hat{S}} = \{(T(\mathbf{x}_{k}),\mathbf{y}_{k})\}_{k=1}^{m}$.

We compare different sets, with different combinations of
color/geometric transformations and noise injection. As we will detail later, we can also generate arbitrary auxiliary meta-domains in this way.

\myparagraph{Domain randomization}
As mentioned in Section~\ref{sec:intro}, part of our analysis is aimed at showing that domain randomization helps mitigating catastrophic forgetting. We perform it in a simple fashion~\cite{volpi2019model}: given
an annotated sample $(\mathbf{x},\mathbf{y})\sim\mathcal{P}$, before feeding it to the current model
$\theta$ and optimizing the objective in Eq.~\eqref{eq:erm}, we transform the image with a transformation
uniformly sampled from the given set
$T\sim\Psi$, obtaining the sample $(T(\mathbf{x}), \mathbf{y})$.

\subsection{A meta-learning algorithm}

\myparagraph{Motivation}
We aim at formulating a training objective that, at the same time, allows (i) 
learning a task of interest $\mathcal{T}$; (ii) mitigating catastrophic
forgetting when the model is transferred to a different domain; (iii) easing
adaptation to a new domain. 

To achieve the second and the third goals, we follow an approach which borrows from the
meta-learning literature~\cite{finn2017model}; we assume that we have access to
a number of meta-domains, and exploit them to run meta-gradient updates throughout
the training procedure. We enforce the loss associated with both the original
domain (our training set) and the meta-domains 
(described later) to be small in the
points reached in the weight space, both reducing catastrophic forgetting and easing adaptation.

Unfortunately, when dealing with domain $\mathcal{D}_i$ we do not have access to the other domains $\mathcal{D}_k,
k \neq i$, and hence cannot use them as meta-domains. Instead, we propose to
automatically produce these meta-domains using standard image
transformations, as mentioned in Section~\ref{sec:da}.
We detail this process at the end of this section.
Throughout the next paragraph, when dealing with a domain $\mathcal{D}_{i}$, we assume the availability of
different meta-domains $\mathcal{\hat{D}}_{i,j}$, each defined by a
set of samples $\mathcal{\hat{S}}_{i,j}$. To ease notation, unless in potentially 
ambiguous cases, 
we will drop the index $i$, i.e.,  ($\mathcal{\hat{D}}_{j}$, $\mathcal{\hat{S}}_{j}$).

\myparagraph{Optimization problem}
Training neural networks typically involves a number of
gradient descent steps that minimize a loss, see for instance
Eq.~\eqref{eq:erm} for the classification task.

In our training procedure, prior to every gradient descent step associated with the current domain, we simulate an arbitrary number of optimization steps
to minimize the losses 
associated with the given meta-domains.
If we are given $K$ different meta-domains, and run a single gradient descent step on each of them at iteration $t$, we obtain
$K$ different points in the weight 
space, defined as 
$\{\hat{\theta}^t_{j}=\theta^t - \alpha \nabla_{\theta} \mathcal{L}_{\mathcal{T}}(\mathcal{\hat{S}}_{j}; \theta^t)\}_{j=1}^K$,
where \textit{j} indicates the $j^\text{th}$ meta-domain.

Our core idea is to use these weight configurations to compute the loss associated with the primary domain $\mathcal{D}_{i}$ (observed through the provided training set $\mathcal{S}_{i}$) after adaptation,
$\{\mathcal{L}_{\mathcal{T}}(\mathcal{S}_i; \hat{\theta}^t_{j})\}_{j=1}^K$. Minimizing these loss values enforces the model to be less prone to catastrophic forgetting, according to the definition we have provided in Section~\ref{sec:probform}; we define their sum as $\mathcal{L}_{recall}$.  

Furthermore, we compute the loss values associated with samples from the 
meta-domains
$\{\mathcal{L}_{\mathcal{T}}(\mathcal{\hat{S}}_{j}; \hat{\theta}^t_{j})\}_{j=1}^K$, and
define their sum as $\mathcal{L}_{adapt}$. 
If one divides the samples from the meta-domains in meta-train and meta-test
subsets, minimizing $\mathcal{L}_{adapt}$ is equivalent to running
the MAML algorithm~\cite{finn2017model}.
Combining the pieces together, the loss that we minimize at each step is

\vspace{-10pt}
{\small
\begin{equation}\label{eq:reg_loss}
\mathcal{L} \coloneqq \mathcal{L}_{\mathcal{T}}(\mathcal{S}_i; \theta^t)
+ \beta \underbrace{\frac{1}{K} \sum_{j=1}^K \mathcal{L}_{\mathcal{T}}(\mathcal{S}_i; \hat{\theta}^t_{j})}
_{\text{\footnotesize {\color{orange}$\mathcal{L}_{recall}$}}}
+ \gamma \underbrace{\frac{1}{K} \sum_{j=1}^K \mathcal{L}_{\mathcal{T}}(\mathcal{\hat{S}}_{j}; \hat{\theta}^t_{j})}
_{\text{\footnotesize {\color{red}$\mathcal{L}_{adapt}$}}}
\end{equation}
}
\vspace{-10pt}

Intuitively, the three terms of this objective embody the points (i), (ii) and (iii) detailed at
the beginning of this section (learning one task, avoiding catastrophic
forgetting, and encouraging
adaptation, respectively).

In our exposition above, we have assumed that only
a single meta-optimization step is performed for each auxiliary meta-domain. In this case, computing the gradients 
$\nabla_{\theta}\mathcal{L}_{\mathcal{T}}(\hat{\theta}^t_{j})$ involves the computation of the gradient of a
gradient, since $\nabla_{\theta}\mathcal{L}_{\mathcal{T}}(\hat{\theta}^t_{j})$ =
$\nabla_{\theta}\mathcal{L}_{\mathcal{T}}(\theta^t 
- \alpha \nabla_{\theta} \mathcal{L}_{\mathcal{T}}(\theta^t))$~\cite{finn2017model}. One could define multi-step meta-optimization procedures, but we do not cover that extension in this work.

For Eq.~\eqref{eq:reg_loss} to be practical, we need access to meta-domains $\mathcal{\hat{D}}_{j}$
during training---concretely, sample sets $\mathcal{\hat{S}}_{j}$ to run the meta-updates. The core idea is to generate an arbitrary number of ``auxiliary''
meta-domains by modifying data points from the original training set $\mathcal{S}_i$ via data manipulations.
As already mentioned, we rely on standard image transformations introduced in Section~\ref{sec:da} to do
so; in practice, the idea is to start from the original training set $\mathcal{S}_i=
\{(\mathbf{x}_{k},\mathbf{y}_{k})\}_{k=1}^{m}\sim\mathcal{P}_i$, sample image transformations $T_{j}$ from a
given set $\Psi$, $T_{j}\sim\Psi$, and craft auxiliary sets as $\mathcal{\hat{S}}_{j} =
\{(T_{j}(\mathbf{x}_{k}),\mathbf{y}_{k})\}_{k=1}^{m}$---that can be used for meta-learning. The
learning procedure, that we named \emph{\metada}, is detailed in Algorithm~\ref{alg:metatrain}.
In our implementation, we set $K=1$ in Eq.~\eqref{eq:reg_loss}, and approach it via gradient descent---by randomly sampling one different auxiliary transformation prior to each step ($T$ in line 6 being 
the transformation employed to generate the current auxiliary meta-domain).

\input{algo}

\vspace{-10pt}

Again, while for 
clarity we report 
one single gradient descent step for the auxiliary meta-domains in the 
Algorithm box (line 7), the procedure is general and
can be implemented with arbitrary gradient descent trajectories. 
Given a sequence of domains $(D_i)_{i=1}^N$, we run
Algorithm~\ref{alg:metatrain} on each of them sequentially (see Figure~\ref{fig:main}), providing the respective datasets $\mathcal{S}_i$ as input.

%% file: algo.tex
\setlength{\textfloatsep}{5pt}

\begin{algorithm}[H]
\caption{\metada}
\label{alg:metatrain}
\begin{spacing}{1.3}
\begin{algorithmic}[1]
\footnotesize
\State \textbf{Input:}  
training set $\mathcal{S}_i$, auxiliary transformation set $\Psi = \{T_{q}\}_{q=1}^M$,
initial weights $\theta^{i-1}$, hyper-parameters $\eta$ (learning rate), $\alpha$ (meta-learning rate), $\beta$ and $\gamma$
\State \textbf{Output:} weights $\theta^{i}$
\State \textbf{Initialize:} $\theta \gets \theta^{i-1}$
\For{$t=1,...,H$}
\State Sample $(\mathbf{\hat{x}}, \mathbf{\hat{y}})$ uniformly from $\mathcal{S}_i$
\Comment Batch for meta-update
\State Sample $T$ uniformly from $\Psi$
\Comment Random transformation
\State $\hat{\theta}^{t}_T \gets \theta^t - \alpha \nabla_{\theta} 
\mathcal{L}_{\mathcal{T}}(T(\mathbf{\hat{x}}),\mathbf{\hat{y}};\theta^t)$
\Comment Meta-update
\State Sample $(\mathbf{x}, \mathbf{y})$ uniformly from $\mathcal{S}_i$
\Comment Batch for update
\State $\theta^{t+1} \gets \theta^t 
- \eta\nabla_{\theta}\big(\underbrace{ \mathcal{L}_{\mathcal{T}}(\mathbf{x},\mathbf{y};\theta^t)}_{\text{\scriptsize {\color{blue} Current task}}} +$\\
$\phantom{aaaaaaeeeeaa} + \beta \underbrace{ \mathcal{L}_{\mathcal{T}}(\mathbf{x},\mathbf{y};\hat{\theta}^{t}_{T})}_{\text{\scriptsize {\color{orange}Backward transfer}}}
+ \gamma \underbrace{ \mathcal{L}_{\mathcal{T}}(T(\mathbf{x}),\mathbf{y};\hat{\theta}^{t}_{T})\big)}_{\text{\scriptsize {\color{red}Forward transfer}}}$
\EndFor
\State $\theta^{i} \gets \theta^{H+1}$

\end{algorithmic}
\end{spacing}
\end{algorithm}

%% file: 5_experiments.tex
\section{Experiments}\label{sec:exps}

In this section, we first detail our experimental protocols for evaluating lifelong
learning algorithms (Section~\ref{sec:exps_protocol}) in a continual adaptation setting. We detail the different baselines we benchmark against in Section~\ref{sec:exps_baselines}. Finally, we report our results in~Section~\ref{sec:exps_results}.

\subsection{Experimental protocols}
\label{sec:exps_protocol}

\myparagraph{Digit recognition}
We consider standard digit datasets broadly adopted by the computer vision community: MNIST~\cite{lecun1998gradient-basedlearning}, SVHN~\cite{netzer2011reading},
MNIST-M~\cite{ganin2015unsupervised} and SYN~\cite{ganin2015unsupervised}.
To assess
lifelong learning performance, we propose training trajectories in which first we train on samples
from one dataset, then train on samples from a second one, and so on. Given these four datasets, we propose two distinct protocols, defined by the following sequences: MNIST
$\rightarrow$ MNIST-M $\rightarrow$ SYN $\rightarrow$ SVHN and SVHN $\rightarrow$ SYN $\rightarrow$
MNIST-M $\rightarrow$ MNIST, referred to as $P1$ and $P2$, respectively. These allow to assess performance on two different
scenarios, respectively: starting from easy datasets and moving to harder ones, and vice-versa. 

For both protocols, we
use final accuracy on \textit{every} test set as a metric (in \%); each experiment is repeated $n=3$ times and we report averaged results and standard deviations. 
For compatibility, we resize all images to $32\times32$
pixels, and, for each dataset, we use $10,000$ training samples. We use the 
standard PyTorch~\cite{paszke2019pytorch} implementation of ResNet-18~\cite{he15deep}
in both protocols. We train models on each domain for $H=3\cdot10^3$ gradient descent steps, setting the batch size to $64$. We use Adam optimizer~\cite{kingma2014adam} with a learning rate $\eta=3\cdot10^{-4}$, which is re-initialized with $\eta=3\cdot10^{-5}$ after the first domain. For \textit{Meta-DR}, we set $\beta=\gamma=1.0$ and $\alpha=0.1$ (results associated with more hyper-parameters are reported in 
Appendix~\ref{app:res}).
We consider one set with transformations that
only allow for color perturbations ($\Psi_1$), one that
also allows for rotations ($\Psi_2$), and one that also
allows for noise perturbations ($\Psi_3$). 
See Appendix~\ref{app:transf} for more details. 

\myparagraph{PACS}
We consider the PACS dataset~\cite{li2017deeper}, typically used by the domain
generalization community. It comprises images belonging to seven classes,
drawn from four distinct visual domains: Paintings, Photos, Cartoons, and
Sketches. We propose to use this dataset to assess continual learning
capabilities of our models: as in the Digits protocol, we train a model in one
domain, then in a second one, and so on. At the end of the learning sequence, we
assess performance on the test sets of all domains. We consider the sequence
\textit{Sketches} $\rightarrow$ \textit{Cartoons} $\rightarrow$
\textit{Paintings} $\rightarrow$ \textit{Photos} (increasing the level of
realism over time), and repeat each experiment $n=5$ times, reporting mean and standard deviation.
In this
case, we start from an ImageNet~\cite{deng09imagenet} pre-trained model, and
resize images to $224\times 224$ pixels. We rely on SGD optimizer with learning rate $\eta=0.01$, reduced to $\eta=0.001$ after the first domain. We consider a transformation set with color transformations ($\Psi_4$).

\myparagraph{Semantic scene segmentation}
We rely on the Virtual KITTI 2~\cite{cabon2020virtual} dataset to generate
sequences of domains. We use 30 simulated
scenes, each corresponding to one of the 5 different urban city environments and
one of the 6 different weather/daylight conditions (see Figure~\ref{fig:main}). 
Ground-truth for several
tasks is given for each datapoint; here, we focus on the semantic
segmentation task.

As we will show in the next section, the most
severe forgetting happens when the visual conditions change drastically (as
expected); for this reason, we focus on cases where we train an initial model on 
samples from a particular scene, and adapt it to a novel urban environment with a different condition. In 
concrete terms, given three urban environments $A$, $B$, $C$ sampled from the 
five available, we consider the learning sequences Clean $\rightarrow$ Foggy
$\rightarrow$ Cloudy ($P1$), Clean $\rightarrow$ Rainy $\rightarrow$ Foggy
($P2$) and Clean $\rightarrow$ Sunset $\rightarrow$ Morning ($P3$)---where
by ``clean'' we refer to synthetic samples cloned from the original
KITTI~\cite{geiger2012areweready} scenes. For each
protocol, we randomly sample $n=10$ different permutations of environments $A$, $B$,
$C$ and report averaged results and standard deviations (details in Appendix~\ref{app:train}).  

Since Virtual KITTI 2~\cite{cabon2020virtual} does not provide any default
train/validation/test split, for each scene/condition we use the
first $70\%$ of the sequence for training, the next $15\%$ for validation and the final $15\%$ for test.
We use samples from both cameras
and use horizontal mirroring for data augmentation in every experiment. We consider the U-Net~\cite{ronneberger2015unet} architecture with 
a ResNet-34~\cite{he15deep} backbone 
pre-trained on ImageNet~\cite{deng09imagenet}. We train for $20$ epochs on the first sequence, and for $10$ epochs on the following ones, setting the batch size to $8$. We use Adam optimizer~\cite{kingma2014adam} with a learning rate 
$\eta=3\cdot10^{-4}$, which is re-initialized with 
$\eta=3\cdot10^{-5}$ after the first domain.
For \metada{}, we set $\beta=\gamma=10.0$ and $\alpha=0.01$
(results associated with more hyper-parameters are reported in 
Appendix~\ref{app:res}). We consider a transformation set 
that allows for color perturbations (see Appendix~\ref{app:transf}).
We rely on a publicly 
available semantic
segmentation suite~\cite{yakubovskiy2019segmentationmodelspytorch} that is based on PyTorch~\cite{paszke2019pytorch}.
We assess performance on every domain 
explored during the learning trajectory, using mean intersection over union
(mIoU, in [0, 100]) as a metric.

\subsection{Training methods}
\label{sec:exps_baselines}

We are interested in assessing the performance of models trained via domain randomization (\textit{Naive + DR}) and with our proposed meta-learning solution (\metada{}).

First, we benchmark these methods against the \textit{Naive} baseline introduced in
Section~\ref{sec:method}, which simply finetunes the model as new training domains come
along. Then, we consider two \textit{oracle} methods:
If we assume access to every domain at every point in time, we can either train on
samples from the joint distribution from the beginning
($\mathcal{P}_{0} \cup \mathcal{P}_{1} \dots \cup
\mathcal{P}_{T}$, \textit{oracle (all)}), or grow the distribution over iterations
(first train on $\mathcal{P}_{0}$, then on
$\mathcal{P}_{0} \cup \mathcal{P}_{1}$, etc., \textit{oracle (cumulative)}). 
With access to samples from any domain, oracles are not exposed to catastrophic forgetting; 
yet, their performance is not necessarily an upper bound~\cite{lomonaco2017core50}. 

Concerning methods
devised ad hoc for continual
learning, we compare against L2-regularization and EWC approaches~\cite{kirkpatrick2017overcoming}.
For a fair comparison, these algorithms---and the ones below---are implemented
with the same domain randomization strategies used for \metada{} and \textit{Naive + DR}.
We also assess performance when an episodic memory of $M$ samples per encountered domain is stored (by default, $M=100$). Learning procedures do not change, except that at every training step the current domain's batch of samples is stacked with a batch from the memory (Experience Replay, or ER~\cite{chaudhry2019tiny}).
We further benchmark our strategies against the GEM method~\cite{lopexpaz2017nips}, and
using different
values for the memory size. Note that replay-based approaches come with storage overheads,
since the episodic memory grows with the number of domains as $\mathcal{O}(n)$.

\subsection{Results}
\label{sec:exps_results}

\myparagraph{Classification (Digits and PACS)}
We report in Table~\ref{tab:digits_all} and
Table~\ref{tab:pacs_all} the comparison between
models trained on Digits and on PACS, respectively.
\input{table_digits_all}
\begin{figure*}[!th]
	\begin{center}
      \includegraphics[width=1.0\textwidth]{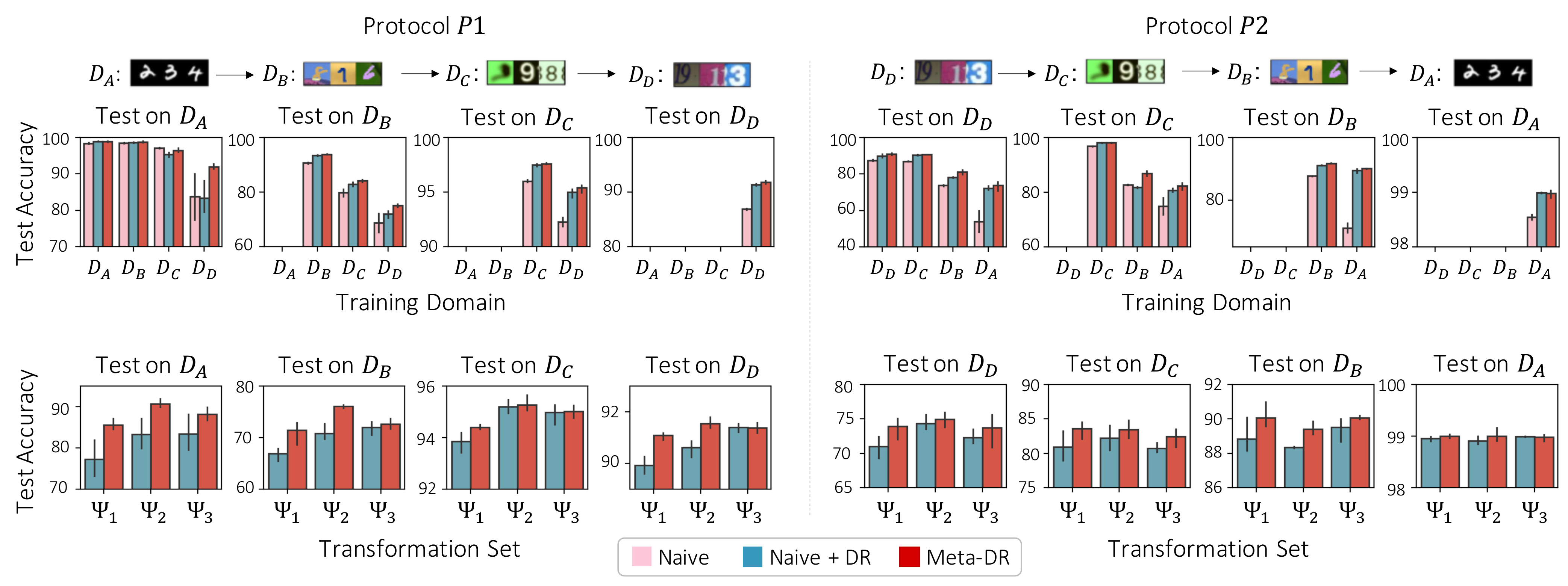}
	\end{center}
	\vspace{-10pt}
	\caption{Digit classification accuracy for protocols $P1$ (left, \ie MNIST $\rightarrow$ MNIST-M $\rightarrow$ SYN $\rightarrow$ SVHN) and $P2$
    (right, \ie SVHN $\rightarrow$ SYN $\rightarrow$ MNIST-M $\rightarrow$ MNIST)  of digit experiments. (Top) Per-domain performance on the test set throughout the
    training sequence (after having trained on each of the four domains). (Bottom)
    Performance at the end of the training sequence for different transformation sets $\Psi_i$. The methods \textit{Naive}, \textit{Naive + DR} and \metada{}, are associated with pink, blue and red bars, respectively.}
\label{fig:digit_results}
\vspace{-10pt}
\end{figure*}

The first result to highlight, is the significantly improved performance of \textit{Naive + DR} with respect to the baseline (\textit{Naive}). For instance, observe the improved performance on SVHN samples at the end of the training sequence of protocol $P2$, from $\sim54\%$ to $\sim72\%$ (Table~\ref{tab:digits_all}); or the improved performance on Sketches, from $\sim74\%$ to $\sim84\%$ (Table~\ref{tab:pacs_all}). This comes without additional overhead on the learning process, except for the neglectable computation spent to transform image samples.

Next, in some cases we can appreciate the benefit of combining domain randomization with methods devised ad hoc for continual learning, namely L2 and EWC~\cite{kirkpatrick2017overcoming}; without data augmentation, we could not detect any significant improvement when using these algorithms in our setting. Note that such regularization strategies generally perform better under the availability of a task label at test time~\cite{maltoni2019continuous,chaudry2018riemannian,farquhar2018robust,kemker2018measuring,hayes2020remind};
in our settings, we desire our model to be able to process samples from any arbitrary domains without further information such as domain labels. 

Our meta-learning approach compares favorably with all non-oracle approaches (see \metada{} row in Tables~\ref{tab:digits_all} and~\ref{tab:pacs_all}). A testbed in which we perform worse than a competing algorithm is SVHN in protocol $P2$, where L2 regularization~\cite{kirkpatrick2017overcoming} performs better. 
SVHN is a more complex domains than the others, so it might be less effective simulating auxiliary meta-domains from this starting point. 

To better highlight the performance gains obtained by relying on meta-learning with respect to the simpler randomization approach and to the naive approach, we report in Figure~\ref{fig:digit_results} (top) the accuracy evolution as the model is transferred and adapted to each of the different
domains, for the two protocols ($P1$ and $P2$, in left and
right panel, respectively). 
\metada{} (red bars) is compared with
\textit{Naive} and \textit{Naive + DR}
(pink and blue bars, respectively). 

To disambiguate the 
contribution of the transformation sets from the contribution of the meta-learning solution, we report in Figure~\ref{fig:digit_results} (bottom)
the performance achieved using different transformation sets $\Psi$ to
generate auxiliary meta-domains. Results are benchmarked against
\textit{Naive + DR} trained with the same sets. These results show that the meta-learning
strategy consistently outperforms such baseline
across several choices for the auxiliary set $\Psi$.

\input{table_pacs_all}

\input{table_digits_ablation_hparams}

\myparagraph{Episodic memory}
We report the results associated with models trained with
an additional memory of old samples in
Tables~\ref{tab:digits_all} and~\ref{tab:pacs_all}
for Digits and PACS experiments, respectively (middle rows). 
As previous work
has shown~\cite{chaudhry2019tiny}, the availability
of a memory significantly mitigates forgetting. 
First, we show that if a memory is
available, the methods \textit{Naive + DR} and \metada{} still positively
contribute to the model performance. Next, we show that such improvement
is consistent as we increase the memory size (Table~\ref{tab:pacs_all}).\footnote{Note
that, in our experiments, the ER algorithm~\cite{chaudhry2019tiny} is the best performing baseline
to compare against---for what concerns replay-based methods.} More results can be found in the Supplementary Material.

\myparagraph{Ablation study}
We report an ablation study in Table~\ref{tab:abl_hparams}, analyzing 
performance of models trained via \metada{} (relying on the set $\Psi_3$) as we include the different terms in our proposed loss (in 
Eq.~\ref{eq:reg_loss}). 
We report performance of
models trained on Digits (protocol $P1$). Accuracy values were computed
after having trained on the four datasets. These results
show that, in
this setting, the first regularizer helps retaining
performance on older tasks (cf. MNIST performance with
and without $\mathcal{L}_{recall}$). Without the second regularizer
though, performance on late tasks suffers (cf.
performance on SYN and SVHN with and without $\mathcal{L}_{adapt}$).
The two
terms together (last row) 
allow for good performance on early tasks as well as
good adaptation to new ones.

\myparagraph{Semantic scene segmentation}
We report in Figure~\ref{tab:semsegm_all} (left) results related with
protocols $P1$, $P2$ and $P3$ (top, middle and bottom, respectively).
\input{table_semsegm_with_fig}
We compare
\metada{} with
\textit{Naive} and \textit{Naive + DR}. 
Also in these settings, heavy data augmentation
proved to be effective to better remember the previous domains;
in general, using \textit{Meta-DR} allows for comparable or better performance than \textit{Naive + DR},
using the same transformation set. We can observe smaller
improvements with respect to the \textit{Naive} baseline when the domain shift is less pronounced ($P3$): in this case, neither domain randomization nor \metada{} carry the same 
benefit that we observed in the other protocols, or in the experiments with different 
benchmarks (Tables~\ref{tab:digits_all} and~\ref{tab:pacs_all}).

%% file: table_digits_all.tex
\begin{table*}\label{tab:abl_sets}
\begin{center}
{
\scriptsize 
\setlength{\tabcolsep}{3.0pt}
\begin{tabular}{@{}lccccccccc@{}}
\multicolumn{10}{c}{\textbf{Digits experiment: comparison}} \\
\toprule

& \multicolumn{4}{c}{\textbf{Protocol $P1$}} 
& & \multicolumn{4}{c}{\textbf{Protocol $P2$}} \\

\cmidrule(r){2-5} 
\cmidrule(r){7-10} 

  \textbf{Method} & \textbf{MNIST (1)} & \textbf{MNIST-M (2)} & ~~ \textbf{SYN (3)} ~~ & ~~ \textbf{SVHN (4)} ~~ & ~~ ~~ & ~~ \textbf{SVHN (1)} ~~ & ~~ \textbf{SYN (2)} ~~ & \textbf{MNIST-M (3)} & \textbf{MNIST (4)} \\

\midrule
\midrule

Naive
& $83.7 \pm 6.4$ & $68.8 \pm 3.4$ & $92.3 \pm 0.4$ & $86.9 \pm 0.1$ &
& $54.0 \pm 5.8$ & $74.9 \pm 3.1$ & $71.1 \pm 1.5$ & $98.5 \pm 0.0$ \\

\midrule

Naive + DR
& $83.4 \pm 3.6$ & $72.0 \pm 1.1$ & $95.0 \pm 0.3$ & $91.4 \pm 0.1$ & 
& $72.3 \pm 0.9$ & $80.8 \pm 0.6$ & $89.5 \pm 0.6$ & $99.0 \pm 0.0$ \\

\midrule

L2~\cite{kirkpatrick2017overcoming} + DR
& $85.9 \pm 2.8$ & $71.8 \pm 1.8$ & $95.4 \pm 0.2$ & $91.4 \pm 0.1$ &
& $75.3 \pm 1.4$ & $82.0 \pm 1.3$ & $89.4 \pm 0.9$ & $98.8 \pm 0.0$ \\

\midrule

EWC~\cite{kirkpatrick2017overcoming} + DR
& $87.2 \pm 1.8$ & $70.7 \pm 1.0$ & $95.4 \pm 0.3$ & $91.8 \pm 0.1$ &
& $73.3 \pm 0.5$ & $80.5 \pm 0.8$ & $89.8 \pm 0.6$ & $98.8 \pm 0.1$ \\

\midrule

\metada
& $92.0 \pm 0.6$ & $75.1 \pm 0.5$ & $95.3 \pm 0.3$ & $91.9 \pm 0.2$ &
& $73.8 \pm 2.1$ & $82.4 \pm 1.1$ & $90.1 \pm 0.1$ & $99.0 \pm 0.1$ \\

\midrule
\midrule

ER~\cite{chaudhry2019tiny}
& $93.2 \pm 0.9$ & $77.7 \pm 1.2$ & $94.1 \pm 0.2$ & $89.2 \pm 0.5$ &
& $74.4 \pm 0.6$ & $86.1 \pm 0.1$ & $89.9 \pm 0.3$ & $98.6 \pm 0.2$ \\

ER~\cite{chaudhry2019tiny} + DR
& $93.9 \pm 0.3$ & $79.9 \pm 0.4$ & $95.8 \pm 0.2$ & $91.5 \pm 0.3$ &
& $80.6 \pm 0.5$ & $90.1 \pm 1.2$ & $89.8 \pm 0.7$ & $98.8 \pm 0.1$ \\

ER~\cite{chaudhry2019tiny} + \metada
& $93.4 \pm 0.8$ & $79.7 \pm 0.4$ & $95.8 \pm 0.5$ & $92.4 \pm 0.1$ & 
& $82.4 \pm 0.4$ & $90.5 \pm 0.2$ & $90.4 \pm 0.1$ & $99.0 \pm 0.1$ \\

\midrule
\midrule

Oracle (all) 
& $99.3 \pm 0.0$ & $93.4 \pm 0.4$ & $97.1 \pm 0.2$ & $89.9 \pm 0.5$ &
& $89.9 \pm 0.5$ & $97.1 \pm 0.2$ & $93.4 \pm 0.4$ & $99.3 \pm 0.0$ \\

\midrule

Oracle (cumul.) ~~  ~~  ~~ 
& $ 99.3 \pm 0.1$ & $93.3 \pm 0.2$ & $96.6 \pm 0.1$ & $88.6 \pm 0.7$ &
& $ 90.2 \pm 0.2$ & $97.0 \pm 0.1$ & $92.5 \pm 0.1$ & $98.5 \pm 0.1$ \\

\bottomrule

\end{tabular}
} 
\end{center}
\caption{Digit classification accuracy on MNIST, MNIST-M, SYN and SVHN at
  the end of protocols $P1$ (left) and $P2$ (right). \metada{} indicates
  results obtained via Algorithm~\ref{alg:metatrain}.
  The same transformation set $\Psi_3$ is used for \metada{} and the methods
  that rely on domain randomization (+ DR).
  Oracles are not comparable as they can access data from all domains at anytime during training. ER-based approaches~\cite{chaudhry2019tiny} rely on an episodic memory, with $100$ samples per domain here.} 
\label{tab:digits_all}
\vspace{-10pt}
\end{table*}

%% file: table_pacs_all.tex
\begin{table}[!t]
\begin{center}
{\scriptsize
\setlength{\tabcolsep}{2.4pt}
\begin{tabular}{@{}lccccc@{}}
 \multicolumn{6}{c}{\textbf{PACS experiment}} \\
\toprule

\textbf{Methods} & \textbf{M.size} & \textbf{Sketches (1)} & \textbf{Cartoons (2)} & \textbf{Paintings (3)} & \textbf{Photos (4)} \\

\midrule
\midrule
Naive& {$-$} & {$73.0 \pm 2.9$} & {$71.4 \pm 3.0$} & {$78.3 \pm 1.6$} & {$98.8 \pm 0.2$}  \\
\midrule

Naive + DR & {$-$} & {$84.3 \pm 2.4$} & {$75.0 \pm 2.2$} & {$80.7 \pm 0.7$} & {$98.7 \pm 0.1$}  \\
\midrule

L2~\cite{kirkpatrick2017overcoming} + DR& {$-$} & {$83.9 \pm 2.3$} & {$73.5 \pm 2.8$} & {$81.5 \pm 1.0$} & {$99.0 \pm 0.2$}  \\
\midrule

EWC~\cite{kirkpatrick2017overcoming} + DR& {$-$} & {$84.8 \pm 1.8$} & {$74.0 \pm 2.1$} & {$80.9 \pm 1.9$} & {$98.6 \pm 0.1$}  \\
\midrule

\metada & {$-$} & {$85.7 \pm 1.8$} & {$75.4 \pm 0.7$} & {$82.0 \pm 1.9$} & {$98.5 \pm 0.3$}  \\

\midrule
\midrule

GEM~\cite{lopexpaz2017nips}
       & {$100$} & {$81.6 \pm 2.9$ } & {$85.4 \pm 0.6$ } & {$82.4 \pm 0.9$ } & {$97.4 \pm 0.2$ }  \\
\rowcolor{gray!10.0}
	  & {$200$} & {$84.3 \pm 0.8$ } & {$86.3 \pm 0.8$ } & {$81.4 \pm 0.6$ } & {$97.2 \pm 0.3$ }  \\
\rowcolor{gray!20.0}
	  & {$300$} & {$82.3 \pm 3.0$ } & {$85.6 \pm 0.9$ } & {$81.6 \pm 0.3$ } & {$96.9 \pm 0.7$ }  \\
\midrule

GEM~\cite{lopexpaz2017nips} + DR 
      & {$100$} & {$88.3 \pm 2.3$ } & {$84.3 \pm 0.2$ } & {$84.9 \pm 0.7$ } & {$97.9 \pm 0.2$ }  \\
	  \rowcolor{gray!10.0}

	  & {$200$} & {$90.0 \pm 0.9$ } & {$85.3 \pm 0.3$ } & {$83.7 \pm 0.2$ } & {$97.6 \pm 0.3$ }  \\
	  \rowcolor{gray!20.0}

	  & {$300$} & {$89.9 \pm 1.2$ } & {$84.1 \pm 0.5$ } & {$85.0 \pm 0.9$ } & {$97.9 \pm 0.2$ }  \\
\midrule

ER~\cite{chaudhry2019tiny}
      & {$100$} & {$88.1 \pm 1.1$ } & {$85.0 \pm 1.7$ } & {$84.1 \pm 1.4$ } & {$98.5 \pm 0.3$ }  \\
	  \rowcolor{gray!10.0}

	  & {$200$} & {$89.5 \pm 0.7$ } & {$85.7 \pm 0.7$ } & {$84.7 \pm 1.5$ } & {$98.6 \pm 0.4$ }  \\
	  \rowcolor{gray!20.0}

	  & {$300$} & {$90.1 \pm 0.6$ } & {$86.5 \pm 0.9$ } & {$85.0 \pm 0.9$ } & {$98.5 \pm 0.3$ }  \\
\midrule

ER~\cite{chaudhry2019tiny} + DR 
      & {$100$} & {$90.9 \pm 0.5$ } & {$85.1 \pm 1.2$ } & {$87.2 \pm 1.0$ } & {$98.7 \pm 0.3$ }  \\
	  \rowcolor{gray!10.0}

	  & {$200$} & {$91.3 \pm 1.0$ } & {$86.3 \pm 0.9$ } & {$87.3 \pm 1.1$ } & {$98.7 \pm 0.2$ }  \\
	  \rowcolor{gray!20.0}

	  & {$300$} & {$91.4 \pm 0.6$ } & {$87.1 \pm 0.4$ } & {$87.4 \pm 1.2$ } & {$98.5 \pm 0.3$ }  \\
\midrule

ER~\cite{chaudhry2019tiny} + \metada 
      & {$100$} & {$91.5 \pm 0.8$ } & {$87.2 \pm 1.0$ } & {$87.6 \pm 1.4$ } & {$98.4 \pm 0.2$ }  \\
	  \rowcolor{gray!10.0}

	  & {$200$} & {$92.5 \pm 0.5$ } & {$87.7 \pm 0.8$ } & {$87.8 \pm 0.8$ } & {$98.7 \pm 0.2$ }  \\
	  \rowcolor{gray!20.0}

	  & {$300$} & {$92.4 \pm 0.6$ } & {$88.1 \pm 0.5$ } & {$88.5 \pm 0.9$ } & {$98.7 \pm 0.3$ }  \\
\midrule
\midrule

Oracle  (all) & {$-$} & {$93.6 \pm 0.3$ } & {$90.7 \pm 0.7$ } & {$88.4 \pm 0.6$ } & {$98.8 \pm 0.2$ }  \\
\midrule
Oracle  (cumul) & {$-$} & {$92.2 \pm 0.7$ } & {$89.1 \pm 1.0$ } & {$86.3 \pm 0.8$ } & {$97.8 \pm 0.4$ }  \\

\bottomrule

\end{tabular}
} 
\end{center}
\caption{Results related to PACS~\cite{li2017deeper}, at the end of the training sequence \textit{Sketches} $\rightarrow$ \textit{Cartoons} $\rightarrow$
\textit{Paintings} $\rightarrow$ \textit{Photos}. GEM~\cite{lopexpaz2017nips} and ER-based approaches~\cite{chaudhry2019tiny} rely on an episodic memory, with a number of samples per domain indicated in the 2\textsuperscript{nd} column.}
\label{tab:pacs_all}
\vspace{-10pt}
\end{table}

%% file: table_digits_ablation_hparams.tex
\begin{table}[!t]
\begin{center}
{\scriptsize
\setlength{\tabcolsep}{2.5pt}
\begin{tabular}{@{}rccccc@{}}
 \multicolumn{6}{c}{\textbf{Digits experiment: ablation study}} \\
\toprule
\multicolumn{2}{c}{\textbf{Losses}} & \multicolumn{4}{c}{\textbf{Protocol: P1}} \\
\cmidrule(r){3-6} 

$\mathcal{L}_{recall}$ & $\mathcal{L}_{adapt}$ 
& \textbf{MNIST (1)} & \textbf{MNIST-M (2)} & \textbf{SYN (3)} & \textbf{SVHN (4)} \\

\midrule

& & $83.7 \pm 6.4$ & $68.8 \pm 3.4$ & $92.3 \pm 0.4$ & $86.9 \pm 0.1$ \\
\midrule

\checkmark & & $94.3 \pm 0.7$ & $76.5 \pm 0.6$ & $94.4 \pm 0.0$ & $89.5 \pm 0.1$ \\
\midrule

& \checkmark & $89.7 \pm 0.5$ & $74.6 \pm 0.1$ & $95.4 \pm 0.1$ & $91.9 \pm 0.0$ \\
\midrule

\checkmark & \checkmark & $92.0 \pm 0.6$ & $75.1 \pm 0.5$ & $95.4 \pm 0.3$ & $91.9 \pm 0.2$ \\

\bottomrule

\end{tabular}
} %
\end{center}
\caption{Ablation study of the loss terms in Eq.~(\ref{eq:reg_loss}). Performance evaluated on all domains at the end of the training sequence $P1$.} 
\label{tab:abl_hparams}
\end{table}

%% file: table_semsegm_with_fig.tex
\begin{figure}[!t]
  \begin{minipage}{0.60\linewidth}
  \vspace{-0.5cm}
    
        \scriptsize
        \renewcommand{\arraystretch}{1.4}

            \setlength{\tabcolsep}{1.0pt}
            \begin{tabular}{@{}rccccc@{}}
             \multicolumn{4}{c}{\textbf{Semantic segmentation experiment}} \\
            \toprule
             
            & \multicolumn{3}{c}{\textbf{Protocol $P1$}} 
            \\
            
            \cmidrule(r){2-4} 
            
            & \textbf{Clean (1)} & \textbf{Foggy (2)} & \textbf{Cloudy (3)} 
            \\

            \midrule
            
            \color{green(pigment)}{Naive}
            & $56.6 \pm 15.1$ & $34.5 \pm 9.7$ & $78.7 \pm 10.1$ 
            \\
            
            \midrule
            
            \color{blue(ncs)}{Naive + DR}
            & $61.9 \pm 8.8$ & $46.1 \pm 8.6$ & $78.7 \pm 8.9$ 
            \\
            \midrule
            
            \color{brickred}{\metada{}}
            & $63.2 \pm 7.8$ & $51.1 \pm 8.1$ & $79.3 \pm 10.3$ 
            \\
            
            \bottomrule
            
            & \multicolumn{3}{c}{\textbf{Protocol $P2$}} & 
            \\
            
            \cmidrule(r){2-4} 
            
            & \textbf{Clean (1)} & \textbf{Rainy (2)} & \textbf{Foggy (3)} &
            \\

            \midrule
            
            \color{green(pigment)}{Naive}
            & $41.3 \pm 13.7$ & $40.3 \pm 12.8$ & $75.3 \pm 19.1$ &
            \\
            
            \midrule
            
            \color{blue(ncs)}{Naive + DR}
            & $59.6 \pm 8.7$ & $53.8 \pm 11.3$ & $75.4 \pm 9.1$ &
            \\
            \midrule
            
            \color{brickred}{\metada{}}
            & $59.8 \pm 8.8$ & $59.0 \pm 10.5$ & $74.8 \pm 09.6$ &
            \\
            
            \bottomrule
            
            & \multicolumn{3}{c}{\textbf{Protocol $P3$}} 
            \\
            
            \cmidrule(r){2-4} 
            & \textbf{Clean (1)} & \textbf{Sunset (2)} & \textbf{Morning (3)} 
            \\

            \midrule
            
            \color{green(pigment)}{Naive}
            & $60.3 \pm 11.5$ & $63.6 \pm 7.7$ & $76.0 \pm 10.0$
            \\
            
            \midrule
            
            \color{blue(ncs)}{Naive + DR}
            & $61.4 \pm 8.1$ & $62.3 \pm 8.1$ & $73.4 \pm 9.9$ 
            \\
            \midrule
            
            \color{brickred}{\metada{}}
            & $62.6 \pm 9.2$ & $61.5 \pm 8.7$ & $74.5 \pm 11.2$ 
            \\
            
            \bottomrule
            
            \end{tabular}

  \end{minipage}
  \begin{minipage}{0.05\linewidth}
  \hspace{0.5cm}
  \end{minipage}
  \begin{minipage}{0.33\linewidth}
    \vspace{1cm}
    \includegraphics[width=\linewidth]{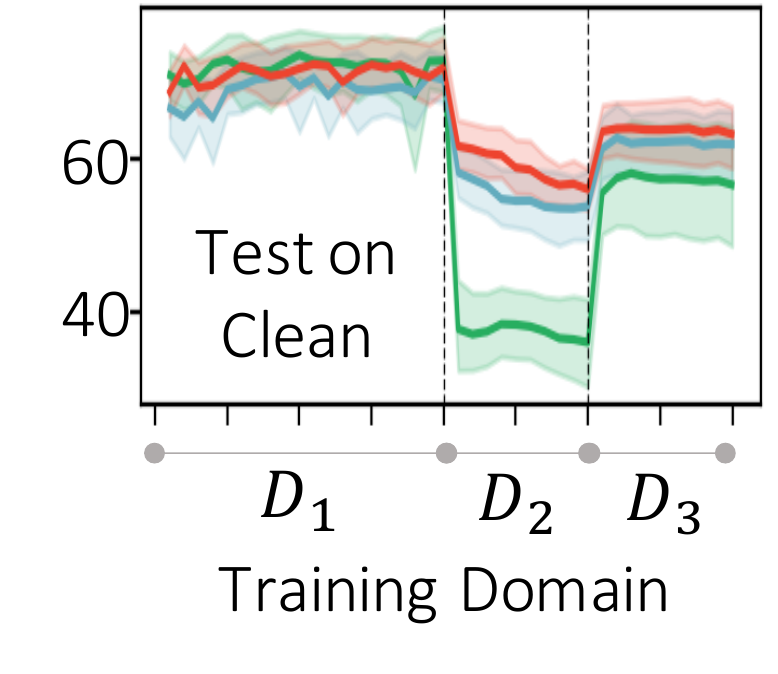}
    \includegraphics[width=\linewidth]{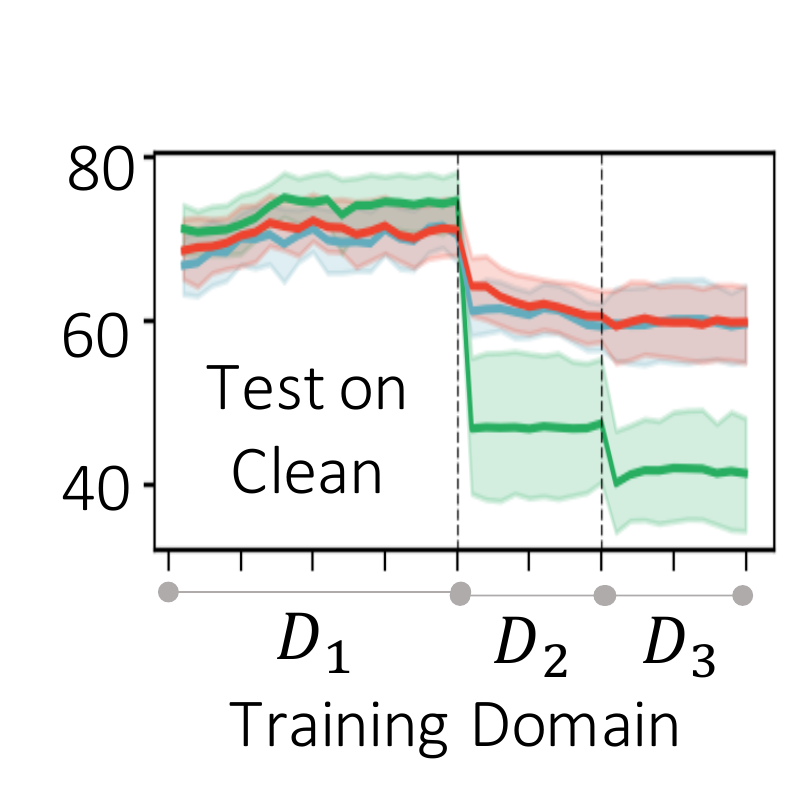}
    \includegraphics[width=\linewidth]{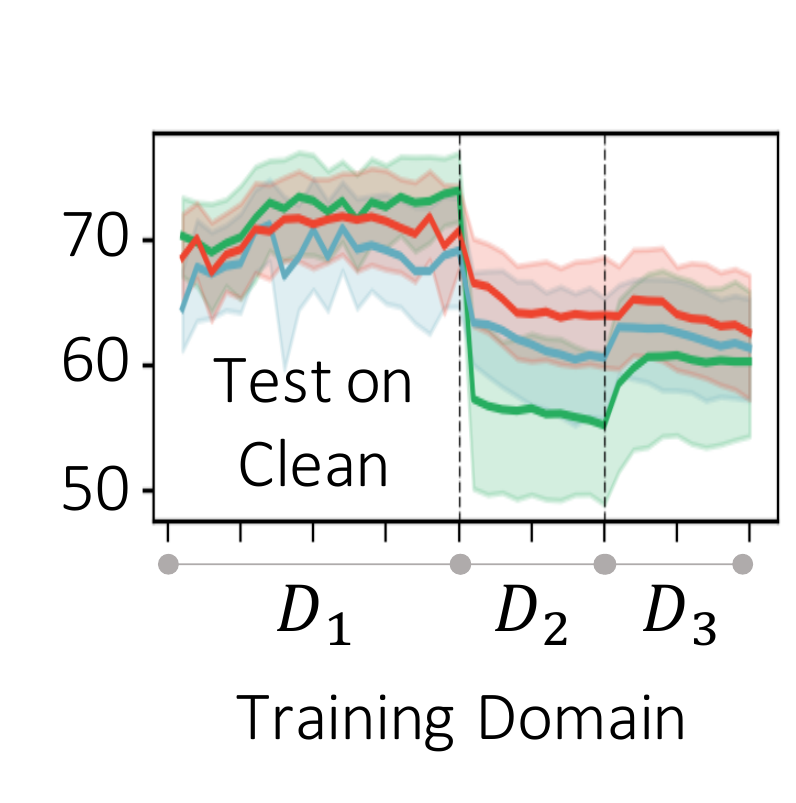}
    
  \end{minipage}
\caption{(Left) mIoUs on domains related to protocols $P1$, $P2$, $P3$ at the end of the training sequences (top, middle and bottom, respectively). (Right) Performance on the first domain ``Clean'' throughout the learning sequences. Legend in Table (Left).}

\label{tab:semsegm_all}
\end{figure}

%% file: 6_conclusions.tex
\vspace{-5pt}
\section{Conclusions and future directions}\label{sec:concl}

We propose and experimentally validate the use of domain randomization in a continual
adaptation setting---more specifically, when a visual
representation needs to be adapted to different domains sequentially. Through our
experimental analysis, we show that heavy data augmentation plays a key role in mitigating
catastrophic forgetting. Apart from the classical way of performing augmentation, we
propose a more effective meta-learning approach based on the concept of ``auxiliary''
meta-domains, that we believe the broader meta-learning field can draw inspiration from. 

For future work, we believe that designing more effective auxiliary sets might significantly improve adaptation to more
diverse domains. For example, previous works focused on augmentation strategies to improve
in-domain~\cite{cubuk2019autoaugment,hu2019learning} and
out-of-domain~\cite{volpi2018generalizing,volpi2019model,qiao2020learning} performance might prove helpful in our settings.

%% file: supplementary.tex
\clearpage
\newpage
{\Large
\textbf{Supplementary Material}
}
\section{Details of the experimental protocol}\label{app:train}

\paragraph{Experiments on the KITTI dataset.}

We have discussed in Section~\ref{sec:exps_protocol} of the main paper three different protocols ($P1$, $P2$, $P3$) to evaluate lifelong learning. Each protocol corresponds to a sequence of conditions (e.g. Clean$\rightarrow$Foggy$\rightarrow$Cloudy for P1) and uses a different urban environment sequence for each condition, which we refer to as A, B, and C in the paper. For each protocol, we train models on 11 different permutations of A, B and C, which we list below for reproducibility (following KITTI's notation~\cite{cabon2020virtual}), and report mean and std results.

\begin{enumerate}
    \item Scene-02 $\rightarrow$ Scene-01 $\rightarrow$ Scene-06
    \item Scene-06 $\rightarrow$ Scene-01 $\rightarrow$ Scene-18  
    \item Scene-20 $\rightarrow$ Scene-01 $\rightarrow$ Scene-18
    \item Scene-02 $\rightarrow$ Scene-18 $\rightarrow$ Scene-20
    \item Scene-06 $\rightarrow$ Scene-01 $\rightarrow$ Scene-02
    \item Scene-20 $\rightarrow$ Scene-18 $\rightarrow$ Scene-01
    \item Scene-02 $\rightarrow$ Scene-06 $\rightarrow$ Scene-01
    \item Scene-18 $\rightarrow$ Scene-20 $\rightarrow$ Scene-02
    \item Scene-20 $\rightarrow$ Scene-06 $\rightarrow$ Scene-01
    \item Scene-18 $\rightarrow$ Scene-06 $\rightarrow$ Scene-02 
    \item Scene-06 $\rightarrow$ Scene-20 $\rightarrow$ Scene-18
\end{enumerate}

\section{Transformation sets used for the auxiliary meta-domains}\label{app:transf}

We report in the Table~\ref{tab:suppl_transf_set} of this supplementary how the transformation sets used for our experiments in Section~\ref{sec:exps} of the main paper are built. We indicate as $\Psi_1$, $\Psi_2$, and $\Psi_3$ the sets used for the digits/PACS experiments (as in Section~\ref{sec:exps_protocol}), and as $\Psi_4$ the set used for the semantic segmentation experiments on KITTI. For the description of a single transformation, we refer to the documentation of the PIL library~\cite{pillow-library} which is the one we used (see in particular~\cite{pillow-image-enhance,pillow-image-ops})---with the exception of \textit{Invert}, \textit{Gaussian noise} and \textit{RGB-rand}. For these three last transformations, we give their details below. Given an RGB image $\mathbf{x}$ with pixels in range $[0,255]$:
\begin{itemize}
    \item \textbf{Invert} applies the transformation $\mathbf{\hat{x}} = |\mathbf{x} - 255|$. 
    \item \textbf{Gaussian noise} perturbs pixels with values that are sampled from a Gaussian distribution with standard deviation $\sigma$ defined by the chosen level. 
    \item \textbf{RGB-rand} perturbs the pixels of each channel by adding factors $r,g,b$, each sampled from a uniform distribution defined in [-level, +level]. 
\end{itemize}

\section{Domain randomization improves domain generalization performance}
\begin{figure*}[h]
	\begin{center}
      \includegraphics[width=\textwidth]{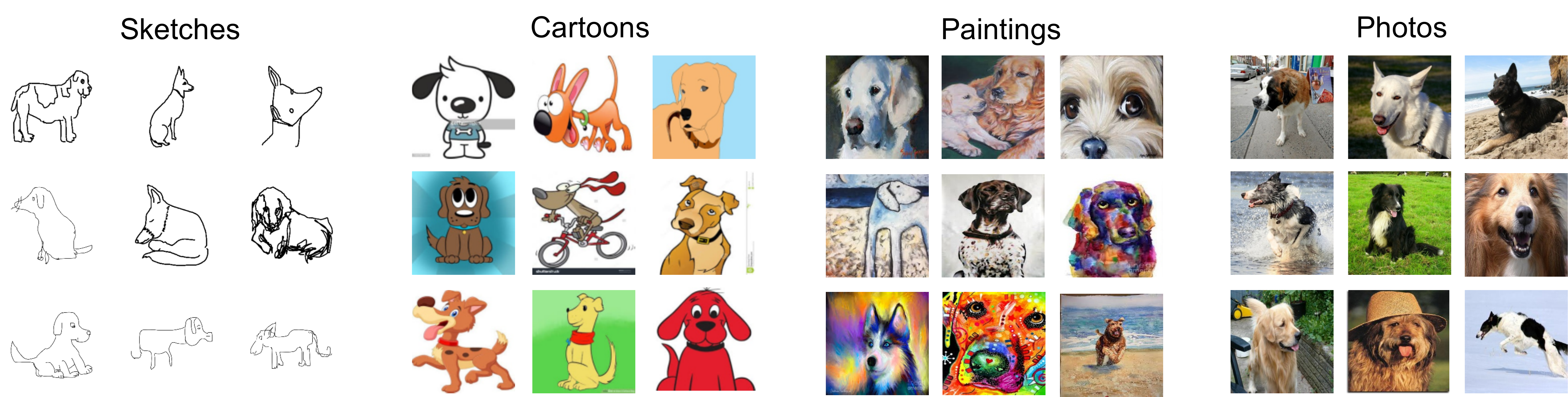}
	\end{center}
	\vspace{-10pt}
	\caption{Samples from the `dog' class of PACS dataset~\cite{li2017deeper}}
\label{fig:pacs_dataset}
\end{figure*}

In Section~\ref{sec:intro} we presented domain randomization as a means to
increase robustness of the model at hand in out-of-domain contexts---and, in
turn, lighten the adaptation process and mitigating the catastrophic forgetting.
We report in Table~\ref{tab:suppl_digits_gen} and~\ref{tab:suppl_pacs_gen} of this supplementary
respectively out-of-domain performance of models trained on MNIST~\cite{lecun2010mnist} and on the
Sketch domain (from
PACS~\cite{li2017deeper}, see Figure~\ref{fig:pacs_dataset}), with and without domain randomization
(relying on transformation set $\Psi_2$ when using domain randomization). 
Similar results for
digits were also shown in previous work~\cite{volpi2019model}. We would like to stress that this protocol is different from the ones used to carry out the experiments in the main manuscript; we are not assessing continual learning performance in this Appendix, but out-of-domain performance of models trained on a single domain (MNIST~\cite{lecun2010mnist} and Sketches~\cite{li2017deeper}). This experiment only serves as a support to our motivation for using domain randomization, expressed in Section~\ref{sec:intro}.

\input{suppl_table_digits_gen}
\input{suppl_table_pacs_gen}

\section{Additional experiments}\label{app:res}
We report in Tables~\ref{tab:supplt_digits_beta},~\ref{tab:supplt_digits_gamma}, and~\ref{tab:supplt_digits_alpha} additional results associated with protocol $P1$ of the digits experiments. We report in Table~\ref{tab:supplt_semsegm_beta} additional results associated with protocol $P3$ of the semantic segmentation experiment on KITTI. All results in Tables~\ref{tab:supplt_digits_beta}--~\ref{tab:supplt_semsegm_beta} are referred to the \metada{} method.

\input{suppl_table_digits_beta}
\input{suppl_table_digits_gamma}
\input{suppl_table_digits_alpha}
\input{suppl_table_semsegm_beta}

\newpage
\phantom{.}
\newpage

We extend the results reported in Table~\ref{tab:digits_all} in the main manuscript
by testing different values for the memory size and further comparison against GEM~\cite{lopexpaz2017nips}; these are reported in Table~\ref{tab:suppl_digits_memory_SGD}, for the protocol $P1$. Note that all methods were implemented with SGD optimizer here (learning rate $\eta=0.01$), for comparability. We further report in Table~\ref{tab:suppl_digits_perms_SGD} results obtained by averaging over the $24$ possible
digit permutations.

\input{suppl_table_digits_memory}

\input{suppl_table_digits_perms}

\input{suppl_transformation_set}

%% file: suppl_table_digits_gen.tex
\begin{table}[h]

\begin{center}
{\small
\setlength{\tabcolsep}{2.5pt}
\begin{tabular}{@{}rccc@{}}
  \multicolumn{4}{c}{\footnotesize \textbf{Domain generalization MNIST models}} \\
\toprule

& \textbf{MNIST-M} & \textbf{SYN} & \textbf{SVHN} \\

\midrule

w/o DR & $41.2 \pm 1.3$ & $35.1 \pm 0.6$ & $23.5 \pm 1.6$ \\
\midrule

w/ DR & $65.6 \pm 5.1$ & $53.7 \pm 2.4$ & $40.4 \pm 1.3$ \\
\bottomrule

\end{tabular}
} 
\end{center}
\caption{\footnotesize Performance of models trained on MNIST~\cite{lecun2010mnist} when tested on MNIST-M~\cite{ganin2015unsupervised}, SYN~\cite{ganin2015unsupervised} and SVHN~\cite{netzer2011reading}. First and second row report results of models trained without and with domain randomization, respectively. These results are related to models trained on a single domain, hence they are not comparable with the ones from the main manuscript.} 
\label{tab:suppl_digits_gen}
\end{table}

%% file: suppl_table_pacs_gen.tex
\begin{table}[h]

\begin{center}
{\small
\setlength{\tabcolsep}{2.5pt}
\begin{tabular}{@{}rccc@{}}
 \multicolumn{4}{c}{\footnotesize \textbf{Domain generalization Sketches models}} \\
\toprule

& \textbf{Cartoons} & \textbf{Paintings} & \textbf{Photos} \\

\midrule

w/o DR & $31.3 \pm 3.0$ & $24.4 \pm 4.3$ & $31.1 \pm 4.3$ \\
\midrule

w/ DR & $48.3 \pm 4.6$ & $28.5 \pm 7.7$ & $36.8 \pm 5.7$ \\
\bottomrule

\end{tabular}
} 
\end{center}
\caption{\footnotesize Performance of models trained on the Sketches domain when tested on Cartoons, Paintings and Photos domains (from PACS~\cite{li2017deeper}). First and second row report results of models trained without and with domain randomization, respectively. These results are related to models trained on a single domain, hence they not comparable with the ones from the main manuscript.} 
\label{tab:suppl_pacs_gen}
\end{table}

%% file: suppl_table_digits_beta.tex
\begin{table}[h]

\begin{center}
{\scriptsize
\setlength{\tabcolsep}{2.5pt}
\begin{tabular}{@{}rcccc@{}}
 \multicolumn{5}{c}{\footnotesize \textbf{Digits experiment: hyper-parameter $\mathbf{\beta}$ }} \\
\toprule
 & \multicolumn{4}{c}{\textbf{Training Protocol: P1}} \\
\cmidrule(r){2-5} 

& \textbf{MNIST (1)} & \textbf{MNIST-M (2)} & \textbf{SYN (3)} & \textbf{SVHN (4)} \\

\midrule

$\beta=0.0$ & $83.7 \pm 6.4$ & $68.8 \pm 3.4$ & $92.3 \pm 0.4$ & $86.9 \pm 0.1$ \\
\midrule

$\beta=0.1$ & $90.6 \pm 2.5$ & $73.7 \pm 1.6$ & $93.6 \pm 0.1$ & $87.9 \pm 0.0$ \\
\midrule

$\beta=1.0$ & $94.3 \pm 0.7$ & $76.5 \pm 0.6$ & $94.4 \pm 0.0$ & $89.5 \pm 0.2$ \\

\bottomrule

\end{tabular}
}
\end{center}
\caption{\footnotesize Performance of models trained with \metada{} with different values for $\beta$ ($\gamma=0.0$). Results averaged over $3$ runs, and models trained using $\Psi_3$. Performance evaluated on all domains at the end of the training sequence $P1$.} 
\label{tab:supplt_digits_beta}
\end{table}

%% file: suppl_table_digits_gamma.tex
\begin{table}[h]

\begin{center}
{\scriptsize
\setlength{\tabcolsep}{2.5pt}
\begin{tabular}{@{}rcccc@{}}
 \multicolumn{5}{c}{\footnotesize \textbf{Digits experiment: hyper-parameter $\mathbf{\gamma}$ }} \\
\toprule
 & \multicolumn{4}{c}{\textbf{Training Protocol: P1}} \\
\cmidrule(r){2-5} 

& \textbf{MNIST (1)} & \textbf{MNIST-M (2)} & \textbf{SYN (3)} & \textbf{SVHN (4)} \\

\midrule

$\gamma=0.0$ & $83.7 \pm 6.4$ & $68.8 \pm 3.4$ & $92.3 \pm 0.4$ & $86.9 \pm 0.1$ \\
\midrule

$\gamma=0.1$ & $91.5 \pm 1.3$ & $76.5 \pm 0.7$ & $94.8 \pm 0.3$ & $89.7 \pm 0.5$ \\
\midrule

$\gamma=1.0$ & $89.7 \pm 0.5$ & $74.6 \pm 0.1$ & $95.4 \pm 0.1$ & $91.9 \pm 0.0$ \\

\bottomrule

\end{tabular}
} 
\end{center}
\caption{\footnotesize Performance of models trained with \metada{} with different values for $\gamma$ ($\beta=0.0$). Results averaged over $3$ runs, and models trained using $\Psi_3$. Performance evaluated on all domains at the end of the training sequence $P1$.} 
\label{tab:supplt_digits_gamma}
\end{table}

%% file: suppl_table_digits_alpha.tex
\begin{table}[h]

\begin{center}
{\scriptsize
\setlength{\tabcolsep}{2.5pt}
\begin{tabular}{@{}rcccc@{}}
 \multicolumn{5}{c}{\footnotesize \textbf{Digits experiment: hyper-parameter $\mathbf{\alpha}$}} \\
\toprule
 & \multicolumn{4}{c}{\textbf{Training Protocol: P1}} \\
\cmidrule(r){2-5}

& \textbf{MNIST (1)} & \textbf{MNIST-M (2)} & \textbf{SYN (3)} & \textbf{SVHN (4)} \\

\midrule
$\alpha=0.001$ & $85.5 \pm 1.6$ & $70.7 \pm 0.7$ & $94.5 \pm 0.3$ & $91.1 \pm 0.0$ \\

\midrule
$\alpha=0.01$ & $87.1 \pm 1.1$ & $72.7 \pm 0.5$ & $95.1 \pm 0.1$ & $91.5 \pm 0.0$ \\

\midrule
$\alpha=0.1$ & $92.0 \pm 0.6$ & $75.1 \pm 0.5$ & $95.4 \pm 0.3$ & $91.9 \pm 0.2$ \\

\bottomrule

\end{tabular}
} 
\end{center}
\caption{\footnotesize Performance of models trained with \metada{} with different values for the meta-learning rate $\alpha$ ($\beta=\gamma=1.0$). Results averaged over $3$ runs, and models trained using $\Psi_3$. Performance evaluated on all domains at the end of the training sequence $P1$.} 
\label{tab:supplt_digits_alpha}
\end{table}

%% file: suppl_table_semsegm_beta.tex
\begin{table}[h]

\begin{center}
{\scriptsize
\setlength{\tabcolsep}{2.5pt}
\begin{tabular}{@{}rccc@{}}
\multicolumn{4}{c}{\footnotesize \textbf{Sem. segm. experiment: hyper-parameter $\mathbf{\beta}$ }}\\
\toprule

& \multicolumn{3}{c}{\textbf{Training Protocol: P3}} \\
\cmidrule(r){2-4} 

& \textbf{Clone (1)} & \textbf{Sunset (2)} & \textbf{Morning (3)} \\
\midrule

$\beta=0.0$ & $60.3 \pm 11.5$ & $63.6 \pm 7.7$ & $76.0 \pm 10.0$ \\
\midrule

$\beta=0.001$ & $62.3 \pm 9.2$ & $67.1 \pm 8.6$ & $73.8 \pm 9.2$ \\
\midrule

$\beta=0.01$ & $61.7 \pm 9.4$ & $65.8 \pm 7.0$ & $73.8 \pm 9.9$ \\
\midrule

$\beta=0.1$ & $61.6 \pm 11.0$ & $67.1 \pm 7.7$ & $74.9 \pm 8.2$ \\
\midrule

$\beta=1.0$ & $65.4 \pm 5.3$ & $68.1 \pm 3.7$ & $74.5 \pm 3.7$ \\
\midrule

$\beta=10.0$ & $64.1 \pm 7.6$ & $66.6 \pm 6.9$ & $73.8 \pm 8.4$ \\
\bottomrule

\end{tabular}
}
\end{center}
\caption{\footnotesize Performance (mIoU) of models trained with \metada{} with different values for $\beta$ ($\gamma=0.0$). Results averaged over $10$ permutations of urban environments. Performance evaluated on all domains at the end of the training sequence $P3$.} 
\label{tab:supplt_semsegm_beta}
\end{table}

%% file: suppl_table_digits_memory.tex
\begin{table}[h]
\begin{center}
{\scriptsize
\setlength{\tabcolsep}{2.4pt}
\begin{tabular}{@{}lccccc@{}}
 
\multicolumn{6}{c}{\footnotesize \textbf{Digits experiment: memory size}} \\
\toprule

\textbf{Methods} & \textbf{M. size} & \textbf{MNIST(1)} & \textbf{MNIST-M(2)} & \textbf{SYN(3)} & \textbf{SVHN(4)} \\

\specialrule{.4pt}{.4pt}{0pt}

GEM~\cite{lopexpaz2017nips}
      & {$200$} & {$93.77 \pm 0.8$} & {$75.68 \pm 1.1$} & {$93.51 \pm 0.3$} & {$84.58 \pm 1.1$}  \\
	  \rowcolor{gray!12.5}
	  & {300} & {$94.51 \pm 0.7$} & {$76.37 \pm 1.5$} & {$93.68 \pm 0.4$} & {$84.84 \pm 1.1$}  \\
	  \rowcolor{gray!25}
	  & {400} & {$95.19 \pm 0.4$} & {$77.09 \pm 0.9$} & {$93.86 \pm 0.3$} & {$85.16 \pm 0.5$}  \\
\specialrule{.4pt}{.4pt}{0pt}
GEM
+ DR 
      & {$200$} & {$93.59 \pm 0.5$} & {$76.34 \pm 1.2$} & {$95.80 \pm 0.2$} & {$89.82 \pm 0.6$}  \\
	  \rowcolor{gray!12.5}
	  & {300} & {$93.81 \pm 0.7$} & {$77.66 \pm 0.6$} & {$95.65 \pm 0.3$} & {$89.86 \pm 0.6$}  \\
	  \rowcolor{gray!25}
	  & {400} & {$94.23 \pm 0.8$} & {$77.83 \pm 1.2$} & {$95.81 \pm 0.2$} & {$89.96 \pm 0.5$}  \\
\specialrule{.4pt}{.4pt}{0pt}
ER~\cite{chaudhry2019tiny} & {$200$} & {${95.78 \pm 0.3}$} & {$79.88 \pm 0.5$} & {$93.23 \pm 0.2$} & {$86.29 \pm 0.4$}  \\
	  \rowcolor{gray!12.5}
	  & {300} & {${96.41 \pm 0.3}$} & {$81.32 \pm 0.5$} & {$93.50 \pm 0.2$} & {$86.20 \pm 0.4$}  \\
	  \rowcolor{gray!25}
	  & {400} & {$96.63 \pm 0.3$} & {$82.07 \pm 0.5$} & {$93.69 \pm 0.2$} & {$86.43 \pm 0.2$}  \\
\specialrule{.4pt}{.4pt}{0pt}
ER
+ DR 
      & {$200$} & {${95.52 \pm 0.5}$} & {$82.54 \pm 0.7$} & {$95.74 \pm 0.2$} & {$89.96 \pm 0.4$}  \\
	  \rowcolor{gray!12.5}
	  & {300} & {$95.63 \pm 0.4$} & {$84.26 \pm 0.7$} & {$95.94 \pm 0.1$} & {$90.02 \pm 0.3$}  \\
	  \rowcolor{gray!25}
	  & {400} & {$96.45 \pm 0.3$} & {$85.50 \pm 0.3$} & {$95.88 \pm 0.2$} & {$89.94 \pm 0.3$}  \\
\specialrule{.4pt}{.4pt}{0pt}
ER
+ \metada 
      & {$200$} & {${96.05 \pm 0.4}$} & {${84.19 \pm 0.6}$} & {${96.42 \pm 0.1}$} & {${91.46 \pm 0.2}$}  \\
	  \rowcolor{gray!12.5}
	  & {300} & {${96.64 \pm 0.4}$} & {${85.66 \pm 0.4}$} & {${96.56 \pm 0.1}$} & {${91.40 \pm 0.2}$}  \\
	  \rowcolor{gray!25}
	  & {400} & {${97.12 \pm 0.3}$} & {${86.81 \pm 0.3}$} & {${96.73 \pm 0.2}$} & {${91.75 \pm 0.2}$}  \\
\specialrule{.4pt}{.4pt}{0pt}
\end{tabular}
} 
\end{center}
\caption{
\footnotesize Comparison between models trained via GEM~\cite{lopexpaz2017nips} and ER~\cite{chaudhry2019tiny} algorithms, with and wihout DR, and \metada{}. Memory size is varied from $200$ to $400$ samples. For comparability, all models were trained using the SGD optimizer, as performed in the PACS experiments in the main manuscript. For what concerns the episodic memory, the number of samples per domain is indicated in the 2\textsuperscript{nd} column.} 
\label{tab:suppl_digits_memory_SGD}
\end{table}

%% file: suppl_table_digits_perms.tex
\begin{table}[h]
\begin{center}
{\scriptsize
\setlength{\tabcolsep}{2.4pt}
\begin{tabular}{@{}lcccc@{}}

\multicolumn{5}{c}{\footnotesize \textbf{Digits experiment: $\mathbf{24}$ permutations}} \\
\toprule

\textbf{Methods} & \textbf{MNIST} & \textbf{MNIST-M} & \textbf{SYN} & \textbf{SVHN} \\

\toprule

GEM~\cite{lopexpaz2017nips}  & {96.48(2.1)} & {81.53(6.4)} & {90.09(5.5)} & {78.16(5.8)}  \\
\midrule
GEM~\cite{lopexpaz2017nips} + DR  & {96.09(2.7)} & {83.45(7.6)} & {90.86(6.5)} & {83.01(6.1)}  \\
\midrule
ER~\cite{chaudhry2019tiny} & {97.23(1.3)} & {84.65(3.7)} & {92.49(2.5)} & {82.53(2.8)}  \\
\midrule
ER~\cite{chaudhry2019tiny} + DR  & {97.04(1.4)} & {86.31(4.2)} & {94.77(1.9)} & {87.01(2.3)}  \\
\midrule
ER~\cite{chaudhry2019tiny} + \metada & {97.67(1.1)} & {87.94(4.0)} & {95.61(1.6)} & {88.82(2.0)}  \\

\bottomrule
\end{tabular}
} 
\end{center}
\caption{\footnotesize Average results for the 24 possible digit permutations that can be obtained from the set of available domains \{MNIST, MNIST-M, SYN, SVHN\}. For what concerns the episodic memory, the number of samples per domain is set to $100$.}
\label{tab:suppl_digits_perms_SGD}
\end{table}

%% file: suppl_transformation_set.tex
\begin{table*}[ht]
\begin{center}
{\normalsize
\setlength{\tabcolsep}{5pt}
\begin{tabular}{@{}rccccccc@{}}
 \multicolumn{7}{c}{\textbf{Image transformations (for auxiliary meta-domains or data augmentation)}} \\
\toprule
& & & \multicolumn{4}{c}{\textbf{Set $\Psi$}} \\
\cmidrule(r){4-7}
\textbf{Transformations} & \textbf{Range} & \textbf{No. Levels} 
& $\mathbf{\Psi_1}$ & $\mathbf{\Psi_2}$ & $\mathbf{\Psi_3}$ & $\mathbf{\Psi_4}$ \\
\midrule
\textit{Brightness}     & $[0.2,1.8]$  & $90$ & \checkmark & \checkmark & \checkmark & \checkmark \\
\midrule
\textit{Color}          & $[0.2,1.8]$  & $90$ & \checkmark & \checkmark & \checkmark & \checkmark \\
\midrule
\textit{Contrast}       & $[0.2,1.8]$  & $90$ & \checkmark & \checkmark & \checkmark & \checkmark \\
\midrule
\textit{RGB-rand}       & $[1,120]$    & $90$ &  & & & \checkmark \\
\midrule
\textit{Solarize}       & $[255,75]$   & $90$ & \checkmark & \checkmark & \checkmark &            \\
\midrule
\textit{Grayscale}      & $-$          & $1$  & \checkmark & \checkmark & \checkmark &            \\
\midrule
\textit{Invert}         & $-$          & $1$  & \checkmark & \checkmark & \checkmark &            \\
\midrule
\textit{Rotate}         & $[-60,60]$   & $30$ &  & \checkmark & \checkmark &            \\
\midrule
\textit{Gaussian noise} & $[0.0,30.0]$ & $30$ &  &  & \checkmark &            \\
\midrule
\textit{Blur}           & $-$          & $1$  &  &  & \checkmark &            \\
\midrule
\\
\midrule
\textbf{Number of transformations $N$} & & & 2 & 2 & 2 & 2\\
\bottomrule

\end{tabular}
} 
\end{center}
\caption{\footnotesize Details of the different transformation sets applied to images, which are either used to create the auxiliary meta-domains or for data augmentation.}
\label{tab:suppl_transf_set}
\end{table*}